\documentclass[10pt,journal,compsoc]{IEEEtran}
\usepackage{amssymb}
\usepackage{amsmath}
\usepackage{array}
\usepackage{multirow}
\usepackage{multicol}
\usepackage{graphicx}
\usepackage{subfigure}
\usepackage{booktabs}

\ifCLASSOPTIONcompsoc
  \usepackage[nocompress]{cite}
\else
  \usepackage{cite}
\fi
\ifCLASSINFOpdf
\else
\fi
\begin{document}
%
\title{Stacked Deconvolutional Network for Semantic Segmentation}
%
%
%
%
\author{
    Jun Fu,
    Jing Liu,~\IEEEmembership{Member,~IEEE},
    Yuhang Wang,
    and Hanqing Lu,~\IEEEmembership{Senior~Member,~IEEE}

\IEEEcompsocitemizethanks{\IEEEcompsocthanksitem J. Fu, J. Liu (corresponding author), Y. Wang and H. Lu are with the National Laboratory of Pattern Recognition, Institute of Automation, Chinese Academy of Sciences and University of Chinese Academy of Sciences, Beijing 100190, China. \protect\\
    E-mail: \{jun.fu,jliu,yuhang.wang,luhq\}@nlpr.ia.ac.cn }
}
\IEEEtitleabstractindextext{%
\begin{abstract}

Recent progress in semantic segmentation has been driven by improving the spatial resolution under Fully Convolutional Networks (FCNs). To address this problem, we propose a Stacked Deconvolutional Network (SDN) for semantic segmentation. In SDN, multiple shallow deconvolutional networks, which are called as SDN units, are stacked one by one to integrate contextual information and guarantee the fine recovery of localization information. Meanwhile, inter-unit and intra-unit connections are designed
to assist network training and enhance feature fusion since the connections improve the flow of information and gradient propagation throughout the network. Besides, hierarchical supervision is applied during the upsampling process of each SDN unit, which guarantees the discrimination of feature representations and benefits the network optimization. We carry out comprehensive experiments and achieve the new state-of-the-art results on three datasets, including PASCAL VOC 2012, CamVid, GATECH. In particular, our best model without CRF post-processing achieves an intersection-over-union score of 86.6\% in the test set.

\end{abstract}

}

\maketitle

\IEEEdisplaynontitleabstractindextext

%
\IEEEpeerreviewmaketitle

\IEEEraisesectionheading{\section{Introduction}\label{sec:introduction}}
\IEEEPARstart{S}{emantic} image segmentation has been one of the most important fields in computer vision, which is to predict the category of
individual pixels in an image. Recently, Deep Convolutional Neural Networks (DCNNs) \cite{lecun1998gradient} have strong learning ability to obtain high-level semantic features, and make remarkable advances in computer vision, including image classification \cite{Hinton_CNN,VGG,he2016deep}, object detection \cite{rcnn,faster_rcnn} and keypoint prediction \cite{deeppose,hourglass}. For semantic segmentation tasks, DCNNs based methods mainly utilize the architecture of Full Convolutional Networks (FCNs) \cite{FCN} which usually adopt a certain pretrained classification network and output a probability map per class for an arbitrary-sized input. However, the classification network with downsampling operations sacrifices the spatial resolution of feature maps to obtain the invariance to image transformations. The resolution reduction results in poor object delineation and small spurious regions in segmentation outputs.

Many approaches have been proposed to solve the above problems. One way is to apply dilated convolutions \cite{fullyconnectedcrfs,deeplabv3,deeplabv2,yu2015multi}. This type of solutions are to upsample the filters with different dilation factors, while the number of filter parameters per position remains unchanged. Accordingly, the receptive fields are enlarged and the larger contextual information is captured without losing the spatial resolution. However those methods output a coarse sub-sampling feature maps which still lose details of object delineation.
Moreover, many methods \cite{parsenet,deeplabv2,deeplabv3,PSPNet} incorporate multi-scale or global feature maps to capture contextual information effectively and fit objects at multiple scales.
Although this strategy further improves the receptive field and captures multi-scale information, it still outputs low-resolution feature maps which impede the generation of detailed boundaries.

Another type of methods are to recover the spatial resolution by an upsampling or deconvolutional path \cite{wang2016objectness,segnet,deconvnet,bayes-segnet}, which can generate high-resolution feature maps for dense prediction. By learning the upsampling process, the low resolution of the feature maps can be restored to the input resolution for pixel-wise classification, which is useful for accurate boundary localization. In the above upsampling solutions, the deconvolutional and unpooling layers are appended with a symmetric structure of the corresponding convolutional and pooling layers. Consequently, the parameter size of the new network is twice as large as the original convolutional structure. Furthermore, the deconvolutional networks are usually built by simply stacking layers, leading to the degradation problem when the depth increases \cite{he2016deep}. To make the model easy to convergence, Wang et al. \cite{wang2016objectness} use the VGG16 network \cite{VGG} as pretrained weights to obtain better initial parameters of deconvolutional network, and Noh et al. \cite{deconvnet} use a two-stage training strategy on single object images and multi-object images, respectively. Due to the difficulties on network optimization, neither of the previous networks can be extended directly to the deeper models, e.g., RestNet \cite{he2016deep} and DenseNet \cite{huang2016densely}, resulting in their limited learning ability.

\begin{figure*}[!t]
  \centering
  \centerline{\includegraphics[width = 1\linewidth]{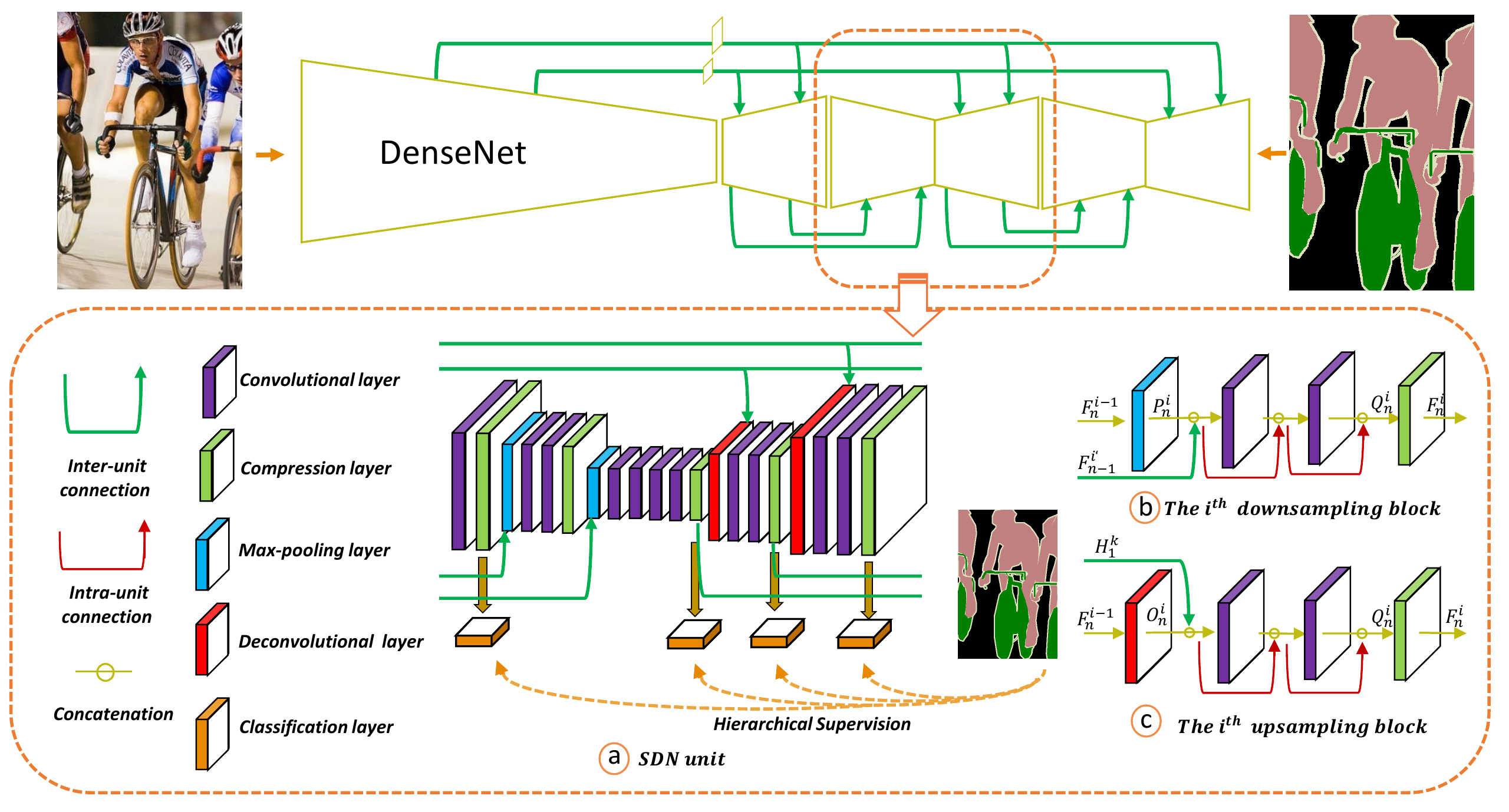}}
  \caption{Overall architecture of our approach. The upper part indicates the structure of proposed stacked deconvolutional network (SDN), the lower part indicates the detailed structure of the SDN unit (a), the downsampling block (b), the upsampling block (c). (Best viewed in color.)}
  \label{SDN}
\end{figure*}

To overcome the above issues, in this paper, we propose a Stacked Deconvolutional Network (SDN) for semantic image segmentation. SDN is a deeper deconvolutional network but easier to optimize compared with most deconvolutional solutions. Instead of building a single deep encoder-decoder network, we design an efficient shallow deconvolutional network (called as SDN unit), and stack multiple SDN units one by one with dense connections, thus making the proposed SDN capture more contextual information and easily optimized. The designing details of the proposed architecture are shown in Fig. \ref{SDN}. A SDN unit is an encoder-decoder network. The encoder module is operated as a downsampling process, aiming to exploit multi-scale features and capture contextual information as much as possible, while the decoder module is operated as an upsampling process to recover the spatial resolution, aiming for accurate boundary localization. To benefit from the impressive performance of the pretrained deep model on ImageNet \cite{ILSVRC}, we adopt DenseNet-161 \cite{huang2016densely} without its last downsampling operations as the encoder module of the first SDN unit, while other encoder and decoder modules consist of some simple downsampling blocks and upsampling blocks, respectively. Typically, each downsampling (upsampling) block includes a max-pooling layer (deconvolutional layer), several convolutional layers, and a compression layer.

As the number of the stacked SDN units increases, the difficulty on model training becomes a major problem. Two strategies have been taken to ameliorate the situation. First, hierarchical supervision is added to upsampling blocks of each SDN unit. Specifically, the compression layers are mapped to pixel-wise labeling maps by a classification layer. With this structure, the network could be learned in a more refined way with no loss on discrimination ability. Second, we import intra-unit and inter-unit connections to help the network optimization. The connections are shortcut paths from early layers to later layers, and they are beneficial to the flow of information and gradient propagation throughout the network. With in an upsampling or a downsampling block of a given SDN unit, the intra-unit connections is a short-range dense connection which is the direct link from the inputs of previous convolutional layers to the ones of back convolutional layers. The inter-unit connection is a long-range skip connection between certain two SDN units. Considering the different intentions for information transmission, there are two forms of inter-unit connections. One is to link decoder and encoder modules between any two adjacent SDN units to promote the network optimization. The other is to connect the multi-scale feature maps from the encoder module of the first SDN unit to the corresponding decoder modules of each SDN unit, thus maintaining the low-level details for the high-resolution prediction.
With the above two strategies, our proposed SDN can be trained efficiently and effectively, and achieves impressive performance on three popular benchmarks including PASCAL VOC 2012 dataset \cite{voc}, CamVid dataset \cite{camvid} and GATECH dataset \cite{GATECH}. In particular, a Mean IoU score of 86.6\% without CRF post-processing on PASCAL VOC 2012 test set.

Our main contributions can be summarized as follows:

\begin{itemize}
   \item We propose a novel network (Stacked Deconvolutional Network) for semantic segmentation, which stacks multiple shallow deconvolutional networks to capture multi-scale context, and guarantee the fine recovery of localization information.
  \item Our proposed SDN adopt intra-unit and inter-unit connections to enhance the flow of information and gradients throughout the network. In particular, inter-unit connections make the multi-scale information across different units efficient to reuse.
  \item The hierarchical supervision for each SDN unit results in better optimization of the proposed network, since early layers of the network can obtain more gradient feedback. Besides, it guarantees the discrimination of the feature maps from the upsampling process of each unit.
  \item Extensive experimental evaluations demonstrate that the proposed SDN model achieves the new state-of-the-art performance on all the three benchmarks.
\end{itemize}

The remainder of this paper is organized as follows: In Section II,
we review the related work about DCNNs based methods for semantic image segmentation and scan the basic architecture of DenseNet \cite{huang2016densely}. In Section III, we will
introduce our proposed Stacked Deconvolutional Network,
including the designing details of a single SDN unit, the connections stacking multiple SDN units and the multi-scale hierarchical supervision.
To verify the effectiveness of our work, the experimental evaluations and necessary analysis are presented in Section IV. Finally, we summarize our work in Section V.

%
%
%
%



\section{Related Work}

Recently, DCNNs based methods make great progress in semantic image segmentation, represented by FCNs \cite{FCN}. FCNs apply a fully convolutional structure and bilinear interpolation to realize pixel-wise prediction, which results in rough edges and object vanishing. Following the fully convolutional structure, many works try to alleviate these problems from the following sides.

Some networks import dilated convolutions and reduce down-sampleing operations, which enable context aggregation and retain more spatial information. Deeplab \cite{fullyconnectedcrfs} and DilatedNet \cite{yu2015multi} adopt dilated convolutions to enlarge receptive fields and capture larger contextual information without losing resolution. \cite{Tusample} employs hybrid dilated convolutions to further enlarge receptive fields of the network. Further on, Dai et al. \cite{deformable} proposes a deformable convolution to adjust the receptive fields according to the scale of objects. The above works remove some downsampling operations for increasing the resolution of outputs, which helps to generate detailed object delineation.

Some networks explore multi-scale or global features to capture contextual information for performance improvement. ParseNet \cite{parsenet} employs global pooling operations to extract image-level information. Deeplabv2 \cite{deeplabv2} proposes atrous spatial pyramid pooling (ASPP) to embed contextual information, which consists of parallel dilated convolutions with different dilated rates. PSPNet \cite{PSPNet} designs a pyramid pooling module to collect the effective contextual prior, containing information with different scales. Deeplabv3 \cite{deeplabv3} proposes an augment ASPP module with image-level features to further capture global context.

Besides, some networks employ deconvolution operations in upsampling path, and thus they are able to realize high resolution prediction and obtain refined object delineation. OA-Seg \cite{wang2016objectness} and SegNet \cite{segnet} apply unpooling operations to unsample the low-resolution features and learn deconvolutional layers to improve the upsampling process. U-Net \cite{U-net} exploits multi-level features by skip connections in a deconvolutional network. RefineNet \cite{refinenet} further refines coarse semantic features with fine-grained low-level features in a multi-path refined architecture. In human pose estimation tasks, Newell et al. \cite{hourglass} stacks encoder-decoder architecture to capture multi-scale information, where the nearest neighbor interpolation is employed in its upsampling path. DCDN \cite{DCDN}, which is our previous work, stacks many small deconvolutional networks for enhancing the learning ability of the network. However, the performances of the above deconvolutional solutions are limited by their difficulties on model optimization or simple network design.

In this work, we stack multiple shallow deconvolutional networks one by one to improve accurate boundary localization.
In contrary to the work \cite{hourglass}, we adopt pretrained classification network to encode images and deconvolutional layers to generate more refined recovery of the spatial resolution, and explore dense connection structure and hierarchical supervision in our network for easier network optimization.
Meanwhile, we expand our previous work DCDN \cite{DCDN} by redesigning deconvolutional network with intra-unit and inter-unit connections, introducing hierarchical supervision, and inheriting the ImageNet \cite{ILSVRC} pre-trained network. All these designs make our network produce more discriminative features and easy to optimize.

Our proposed SDN imports dense connections to make the stacked deep model easy to optimize. This is inspired by the work of DenseNet \cite{huang2016densely}. Accordingly, we will overview the basic idea of DenseNet in the following.

DenseNet adopts dense connections to avoid the vanishing gradient problem and improve the flow of information, and it mainly consists of dense blocks and transition layers. The input of each convolutional layer within a dense block is the concatenation of all feature outputs of its previous layers at a given resolution. Consider ${x_l}$ is the output of the ${l^{th}}$ layer in a dense block, ${x_l}$  can be computed as follows:
\begin{equation}\label{3}
    {x_l} = {H_l}\left( {\left[ {{x_0},{x_1}, \ldots ,{x_{l - 1}}} \right]} \right)
\end{equation}
where ${[ {{x_0},{x_1}, \ldots ,{x_{l - 1}}}]}$ stands for the concatenation of the feature maps ${{x_0},{x_1}, \ldots ,{x_{l - 1}}}$, and ${x_0}$ is the inputs of the dense block. Meanwhile, ${H_l}$ is defined as a composite function of operations:  BN, ReLU, a $1\times1$ convolution operation followed by BN, ReLU, a $3\times3$ convolution operation. Each dense block is followed by a transition layer, which do convolution and pooling to change the number and the size of feature maps. Finally, a softmax classifier is attached to make prediction.

\section{Our Approach}

\subsection{Overview}

We propose a new framework called as Stacked Deconvolutional Network (SDN), which aims to capture more contextual information and recover high-resolution prediction progressively by stacking multiple shallow deconvolutional networks one by one.

As illustrated in Fig. \ref{SDN}, three (but not limited) deconvolutional networks, called as \emph{SDN units}, are piled up from end to end, and the intra-unit connections and the inter-unit connections are jointly employed. Such a connected structure of the network enables efficient backward gradient propagation, and effective detailed boundaries restoration for full resolution labelling prediction. In order to obtain high-quality semantic representation, we adopt DenseNet-161, pre-trained on ImageNet \cite{ILSVRC}, as the encoder module of the first deconvolutional network.
A typical structure of a SDN unit is shown in Fig. \ref{SDN}(a), which consists of two \emph{downsampling blocks} and two \emph{upsampling blocks}.
Each downsampling (upsampling) block starts with a pooling (deconvolutional) layer, crosses a few of convolutional layers and ends with a compression layer. Besides, the hierarchical supervision for multi-level  feature maps are jointly employed during each upsampling process to guarantee discrimination of the output prediction.
In the inference step, we only use the highest-resolution results of the last unit as the final prediction. In the following subsections, we will elaborate the designing details of each SDN unit, intra-unit and inter-unit connections, and the hierarchical supervision.

\subsection{Deconvolutional Network (SDN unit)}

We design a shallow deconvolutional network referred to as a SDN unit to capture contextual information and refine poor object delineation, whose structure is illustrated in Fig. \ref{SDN}(a). A SDN unit has an encoder module and a corresponding decoder module. In a encoder module, we stack two downsampling blocks to enlarge the receptive fields of the network, resulting in the low resolution of the feature maps. In a decoder module, upsampling blocks are used twice to achieve a more refined reconstruction of the feature maps.

For a given SDN unit, its encoder module takes the outputs of its previous unit as inputs and produces low-resolution feature maps with larger receptive fields. Here, we employ the downsampling blocks twice, resulting in $\frac{1}{16}$ spatial resolution of the input image. The structure of downsampling block is shown in Fig. \ref{SDN}(b). A downsampling block consists of a max-pooling layer and 2 (or more) convolutional layers, and a compression layer.
First, the feature map $F^{i-1}_{n}$ from the ${(i-1)^{th}}$ block is fed into the max-pooling layer in the ${i^{th}}$ downsampling block of the ${n^{th}}$ unit and sub-sampled by a factor of 2 to produce a new feature map $P^{i}_{n}$. Second, behind the max-pooling layer, we cascade 2 convolutional layers with intra-unit connections. Concretely, the input of the convolutional layer is the concatenation of the input and output of its previous convolutional layer. Such a densely connected structure is beneficial to feature reuse, i.e., the multi-scale appearance of objects can be better captured to obtain effective semantic segmentation. The densely connected structure takes the feature map $P^{i}_{n}$ and the output $F^{i'}_{n-1}$ of the ${i'^{th}}$ block in the ${(n-1)^{th}}$ unit as inputs and outputs an feature map $Q^{i}_{n}$, where the ${i'^{th}}$ block is the backward nearest block to the ${i^{th}}$ block in the same resolution.
However, intra-unit connections bring the linear growth in the channel number of the feature maps, resulting in too much GPU memory demanding. Finally, in order to decrease the computational cost and the memory demanding, we employ a compression layer, which performs convolutions with fewer filters to reduce the channel number of the feature map $Q^{i}_{n}$, to generate an feature map $F^{i}_{n}$. In short, the ${i^{th}}$ downsampling block takes two feature maps $F^{i-1}_{n}$ and $F^{i'}_{n-1}$ as its inputs, and outputs an new feature map $F^{i}_{n}$, the operations can be summarized as follows:
\begin{equation}\label{eq:downsampling}
\begin{split}
P^{i}_{n} &= Max(F^{i-1}_{n}),\\
Q^{i}_{n} &= Trans([P^{i}_{n},F^{i'}_{n-1}]),\\
F^{i}_{n} &= Comp(Q^{i}_{n}).
  \end{split}
  \end{equation}
where ${[P^{i}_{n},F^{i'}_{n-1}]}$ stands for the concatenation of the feature maps $P^{i}_{n}$ and $F^{i'}_{n-1}$. $Max$($\cdot$) denotes a max-pooling operation. $Trans$($\cdot$) denotes a transformation function of the densely connected structure, in which two sequences of BN, ReLU, a $3\times3$ convolution, dropout, and concatenation operations are performed.
$Comp$($\cdot$) refers to a $3\times3$ convolution operation.
It should be noted that our encoder module of the first SDN unit employs full convolutional DenseNet-161 to obtain high-semantic features, while the other SDN units adopt the encoder module described above.

In the decoder module, we apply upsampling blocks to progressively upsample feature maps to larger resolution. The upsampling blocks are also used twice to enlarge resolution back to $\frac{1}{4}$ spatial resolution of the input image. An upsampling block consists of a deconvolutional layer and several convolutional layers and a compression layer. As shown in Fig. \ref{SDN}(c), in the ${i^{th}}$ upsampling block, we first apply deconvolutional operation on the output $F^{i-1}_{n}$ of the ${(i-1)^{th}}$ block and produce a high-resolution feature map $O^{i}_{n}$. Then the feature map $O^{i}_{n}$ is concatenated with the feature map $H^{k}_{1}$ from the ${k^{th}}$ block in the encoder module of the first SDN unit, where the ${k^{th}}$ block have the same resolution with $O^{i}_{n}$. Finally, the concatenated feature maps are fed to the subsequent convolutional layers and compression layer, which adopt the similar connection structure as the downsampling block. The output $F^{i}_{n}$ of the ${i^{th}}$ upsampling block can be computed as follow:

\begin{equation}\label{eq:downsampling}
\begin{split}
O^{i}_{n} &= Deconv(F^{i-1}_{n}),\\
Q^{i}_{n} &= Trans([O^{i}_{n},H^{k}_{1}]),\\
F^{i}_{n} &= Comp(Q^{i}_{n}).
  \end{split}
  \end{equation}
where $Deconv$($\cdot$) refers to a deconvolutional operation. For the last unsamlping block of the last SDN unit, we abandon its compression layer for better prediction. Note that, our highest resolution of SDN unit is set to a quarter of input images. One important reason for this design is that we can reduce GPU memory usage of a single SDN unit to stack more units.

Contrary to traditional deconvolutional networks \cite{wang2016objectness,deconvnet,segnet}, which have difficulty in network training and require additional aids, our deconvolutional network achieves great improvements in network training. First, our deconvolutional network is shallow encoder-decoder framework, which only includes two simple downsampling and upsampling blocks. Then, intra-unit connections are performed between convolutional layers, and this enables effective backward propagation of the gradients through a SDN unit. All these design improves end-to-end training of all network blocks.

\subsection{Densely Connecting SDN units}
\label{semdeplossfunc}

To enhance the learning ability of network, we stack some shallow SDN units as introduced in above subsection  into a very deep model. Meanwhile, the inter-unit connections are imported to make the multi-scale information across different units efficient to reuse. As shown in Fig. \ref{SDN}, there are two types of inter-unit skip connections in the proposed framework. One is between any two adjacent SDN units, and the other is a kind of skip connections from the first SDN units to others. In the following, we will introduce them in detail.

The skip connections between any two adjacent SDN units are used to promote the flows of high-level semantic information and improve the optimization of encoder modules. For a given SDN unit, its encoder module exploit the intermediate features of the decoder module in its previous unit by a shortcut path.
Specifically, the feature map $P^{i}_{n}$ from the ${i^{th}}$ block of the ${n^{th}}$ unit is concatenated with the feature map $F^{i'}_{n-1}$ from the ${i'^{th}}$ block of the ${(n-1)^{th}}$ unit.
Such a concatenation operation is shown in Fig. \ref{SDN}(b). We adopt the skip connection twice according to the number of downsamlpling blocks. With such skip connections, the gradients can be directly propagated to the previous unit, thus promoting the optimization of network. Besides, the fusion of features from adjacent units would efficiently capture multi-scale information.

Meanwhile, we adopt the skip connections from the first SDN unit to others to fuse low-level representations and high-level semantic features, resulting in refined object segmentation edges. As illustrated in Fig. \ref{SDN}, we feed the low-level visual features from the ${k^{th}}$ block of the first encoder module into a convolutional layer, where the convolution operations generate a feature map $H^{k}_{1}$. Then the feature map $H^{k}_{1}$ is concatenated with the outputs of the deconvolutional layer in the corresponding resolution.
Here, we combine the features from the first encoder module with the upsampling features from each decoder module. This is because mid-level representations in first encoder module have more spatial visual information.
Such inter-unit connections contribute to detailed boundaries for high-resolution prediction.

\subsection{Hierarchical Supervision}
\begin{figure}[!t]
  \centering
  \centerline{\includegraphics[width = 1\linewidth]{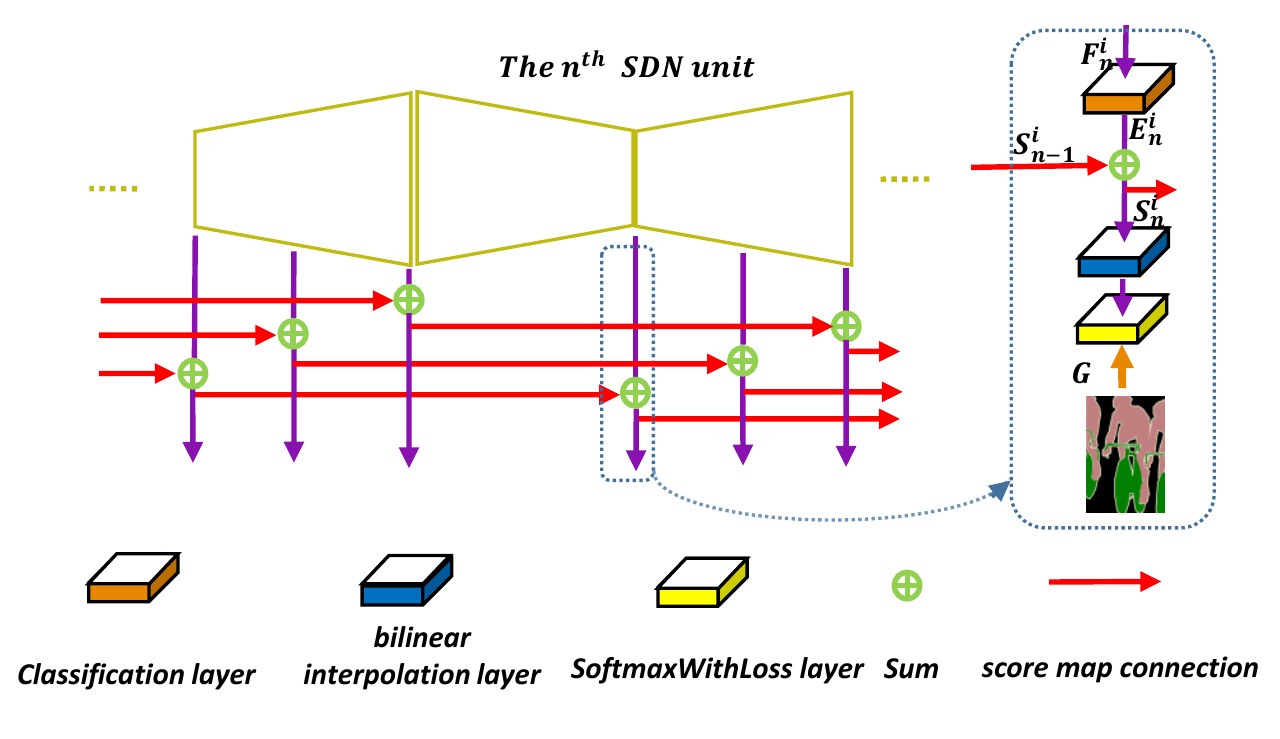}}
  \caption{Hierarchical supervision with score map connections during upsampling process. (Best viewed in color.)}
  \label{Supervision}
\end{figure}
Deeper networks lead to better performance. However, the difficulty in training deeper network becomes a major problem. We stack multiple shallow deconvolutional networks with random initialization, which leads to additional optimization difficulty. According to the previous design, inter-unit and intra-unit connections are used to assist training. In order to alleviate this problems further, we add hierarchical supervision in each SDN unit. Specifically, the output $F^{i}_{n}$ of the ${i^{th}}$ upsampling block is fed to a pixel-wise classification layer to obtain a feature map $E^{i}_{n}$ with channel $C$, where $C$ is the number of possible labels. Then $E^{i}_{n}$ is upsampled to match the size of the input image with bilinear interpolation, and finally supervised with pixel-wise groundtruth. In this way, the hierarchical supervision helps optimize the learning process. The pixel-wise cross-entropy loss is applied to all predictions in the network. For the ${i^{th}}$ block with a pixel-wise cross-entropy loss $\mathcal{L}^{i}$, the loss is computed as follows:

\begin{equation}\label{eq:loss1}
\begin{split}
 \mathcal{L}^{i} = \ell(Bi(E^{i}_{n}),G), E^{i}_{n} = Classify(F^{i}_{n})
  \end{split}
\end{equation}
where $\ell$ denotes a cross-entropy loss function, $Classify$($\cdot$) denotes a classifier with a $3\times3$ convolution operation, $Bi$($\cdot$) denotes a bilinear interpolation operation, and $G$ refers to a ground truth map.

In order to further improve the performance of the network, we enhance score map fusion before bilinear interpolation in the same resolution, depicted in Fig. \ref{Supervision}. In particular, for the ${n^{th}}$ SDN unit, the output $E^{i}_{n}$ of the classification layer are fused with the features $S^{i}_{n-1}$ in the ${(n-1)^{th}}$ unit by element-wise sum operation to produce an new fused feature $S^{i}_{n}$, and then the fused feature $S^{i}_{n}$ is upsampled with bilinear interpolation, and also supervised with pixel-wise groundtruth. Therefor, we can compute the loss $\mathcal{L}^{i}$  as follows:

\begin{equation}\label{eq:loss1}
\begin{split}
\mathcal{L}^{i} &= \ell(Bi(S^{i}_{n}),G),\\
S^{i}_{n} &=  E^{i}_{n} \oplus  S^{i}_{n-1}
  \end{split}
\end{equation}
where $\oplus$ denotes element-wise sum.
Such a same-scale sum fusion of score maps enhances information flows and improves segmentation results.

In the testing phase, we only use the highest-resolution result of the last unit as the final prediction.
It can be found that, with the hierarchical supervision, the feature maps guarantee the discrimination and restrain the noise during upsampling process.

\subsection{Implementation Details}
In each SDN unit, we stack 4 convolutional layers in the block of the lowest resolution for better global perspective, while the other blocks have 2  convolutional layers. The convolutional layers in a block is composed of BN, ReLU, and a $3\times3$ convolution operation followed by dropout with the probability of 0.2. The filter numbers of the convolutional layers are all set to 48. In two downsampling blocks, the compression layers are  $3\times3$ convolution operations and their filter numbers are set to 768 and 1024 respectively. Max-pooling with $2\times2$ window and stride 2 is preformed. Meanwhile, in two upsampling blocks, the compression layers also are $3\times3$ convolution operations and their filter number are set to 768 and 576 respectively. Upsampling operation can be done by a $4\times4$ deconvolutional operation with stride 2. The convolutional layer in inter-unit connections from the first SDN unit to others is composed of BN, ReLU, and a $3\times3$ convolution operation. We employ the full convolutional DenseNet-161 network, pre-trained on ImageNet, as the first encoder module. Meanwhile, the last downsampling operations is removed and dilation convolutions is used.

The proposed SDN is implemented with Caffe \cite{caffe}. Similar to \cite{fullyconnectedcrfs}, we optimize it using the ``ploy" learning rate policy, with batchsize of 10. We set power to 0.9, momentum to 0.9 and weight decay to 0.0005. We apply data augmentation in the training step. Here, random crops of $320\times320$ are used, and horizontal flip is also applied. In the inference step, we pad images with  mean value to make the length divisible by 16 before feeding full images into the network.

\section{Experiments}

To show the effectiveness of our approach, we carry out comprehensive experiments on PASCAL VOC 2012 dataset \cite{voc}, CamVid dataset \cite{camvid} and GATECH dataset \cite{GATECH}. Moreover, we perform a series of ablation evaluations to inspect the impact of various components on PASCAL VOC 2012 dataset. Experimental results demonstrate our proposed SDN achieves new state-of-the-art performance on 3 datasets.

\subsection{Dataset and Evaluation Metrics}

\subsubsection{Dataset}

PASCAL VOC 2012 : The dataset has 1,464 images for training, 1,449 images for validation and 1,456 images for testing, which involves 20 foreground object classes and one background class. Meanwhile, we augment the training set with extra labeled PASCAL VOC images provided by Semantic Boundaries Dataset \cite{SBD}, resulting in 10,582 images for training.

CamVid: The dataset is a street scene understanding dataset which consists of 5 video sequences. Following \cite{segnet}, we split the dataset into 367 training images, 100 validation images, and 233 test images. The resolution of each image is $360 \times 480$ and all images belong to 11 semantic categories. Compared with PASCAL VOC 2012 dataset, CamVid dataset has more strong spatial-relationship among different categories.

GATECH: The dataset is a large video set of outdoor scenes which consists of 63 videos with 12241 frames for training and 38 videos with 7071 frames for testing. The dataset is labeled with 8 semantic classes which are sky, ground, solid, porous, cars, humans, vertical mix, and main mix.

\subsubsection{Evaluation Metrics}
We only report the results of Mean IoU on PASCAL VOC 2012 dataset for the common convention, while we evaluate other datasets with Global Avg and Mean IoU.
\begin{itemize}
    \item Global Avg: Percentage of correctly classified pixels over the whole dataset. i.e. $\frac{\sum_{i}{t_{ii}}}{\sum_{i}{T_i}}$.
    \item Mean IoU: Ratio of correctly classified pixels in a class over the union set of pixels predicted to this class and groundtruth, and then averaged over all classes. i.e. ${\frac{1}{N}}{\sum_{i}{\frac{t_{ii}}{T_{i}+\sum_{j}{t_{ji}}-t_{ii}}}}$.
\end{itemize}
where $N$ is the number of semantic classes, and $T_{i}$ is the total number of pixels in class $i$, while $t_{ij}$ indicates the number of pixels which belong to class $i$ and predicted to class $j$.
\begin{figure}[!t]
  \centering
  \centerline{\includegraphics[width = 1\linewidth]{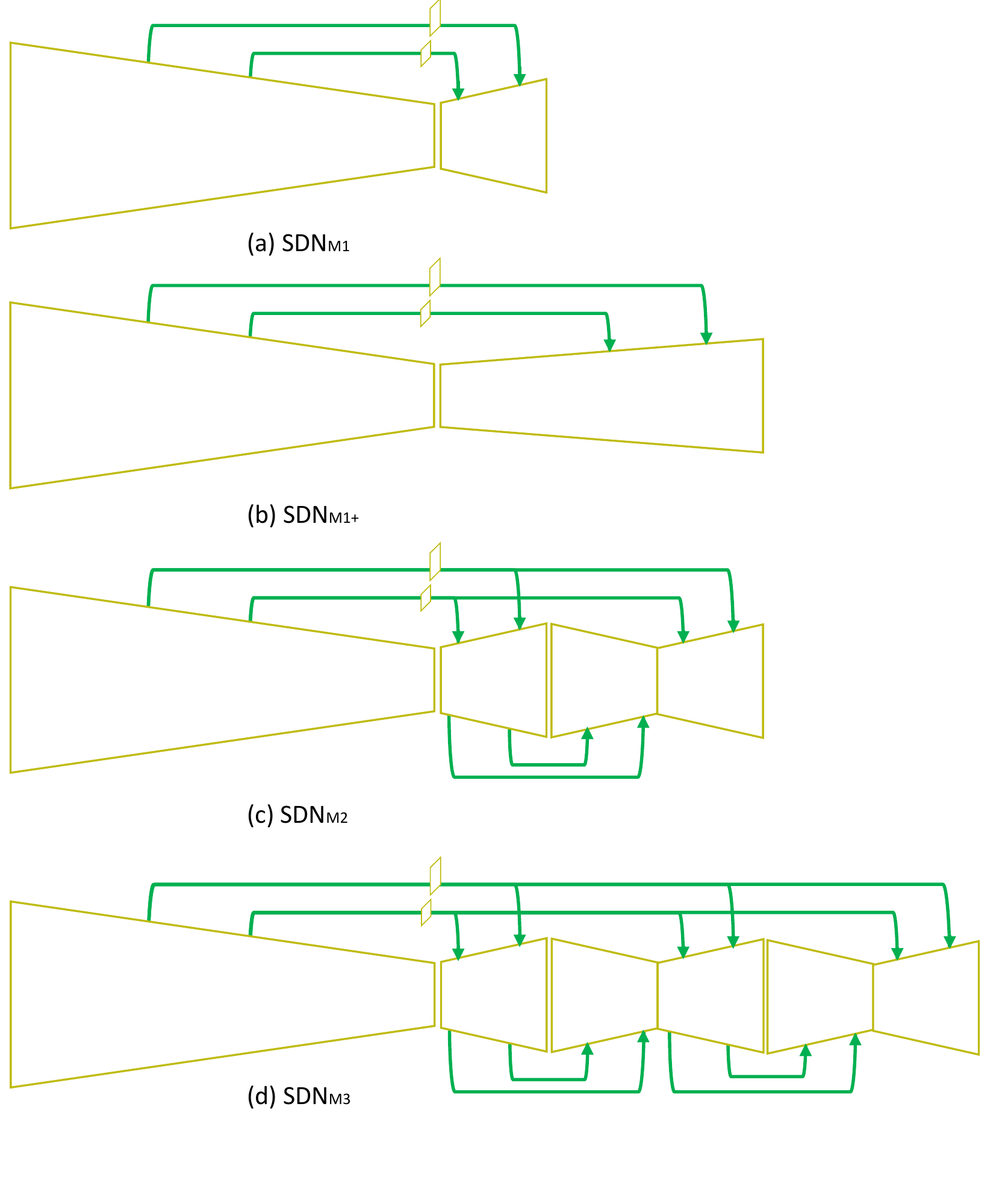}}
  \caption{Different stacked SDN structure. (Best viewed in color.)}
  \label{stack123}
\end{figure}

\subsection{Results on PASCAL VOC 2012 dataset}

In this subsection, we first verify the effectiveness of each component in our approach, including the effects of stacking multiple SDN units, stacked network design and hierarchical supervision. Then we report the experimental results on PASCAL VOC 2012 test set. Here, we set the initial learning rate to 0.00025, and further apply data augmentation by randomly transforming the input images with 5 scales \{0.6, 0.8, 1, 1.2, 1.4\}, and 5 aspect rations \{0.7, 0.85, 1, 1.15, 1.3\}.

\subsubsection{Stacking multiple SDN units}

We stack multiple SDN units to refine the segmentation maps. The key of the stacked architecture is that each unit can capture more contextual information and recover the high-resolution features. To verify this intuition, we gradually increase the number of the SDN units and test the performance respectively. As shown in Fig. \ref{stack123}, we refer to the network stacking $k$ SDN units as SDN${_M{_k}}$, and the number of the SDN units is up to 3.

\begin{figure}[!t]
  \centering
  \centerline{\includegraphics[width = 1\linewidth]{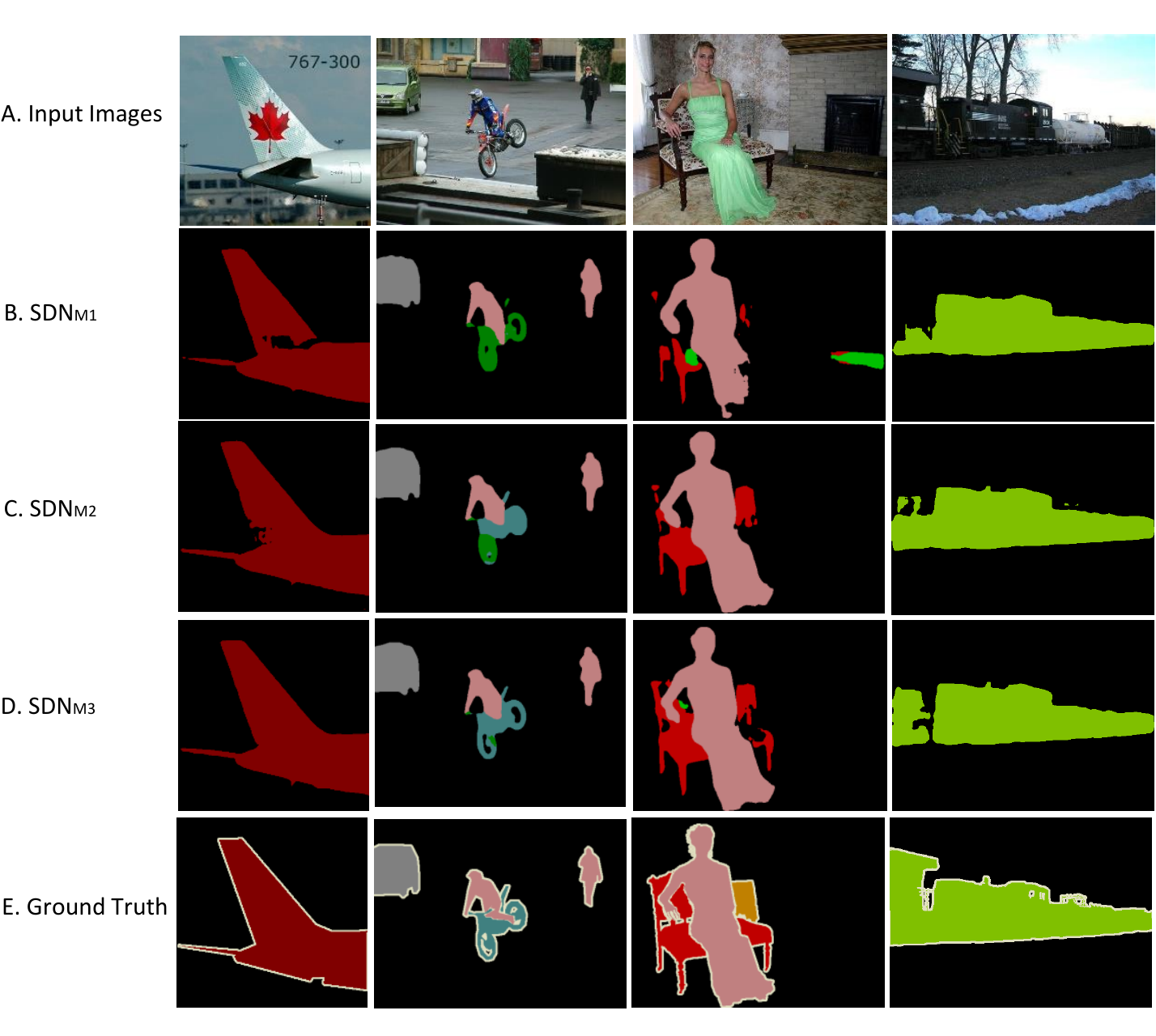}}
  \caption{Results on PASCAL VOC 2012 val set. For every column we list input images (A), the semantic segmentation results of SDN${_M{_1}}$ network (B), SDN${_M{_2}}$ network (C), SDN${_M{_3}}$ network (D), and Ground Truth (E).}
  \label{pascal_picM}
\end{figure}
The results are shown in Table \ref{STACKT}, it is observed that the performance consistently increases with the growth of SDN unit number. Specially, the performance of SDN${_M{_1}}$ network is only 78.2\%. When we increase SDN unit number from 1 to 3, the performance improves from 78.2\% to 79.2\% to 79.9\%. Meanwhile, we show some predicted semantic maps under different networks in Fig. \ref{pascal_picM}. With the growth of SDN unit number, the object edges are ameliorated (the first and fourth column), and the object discriminate is enhanced (the second and third column). These factors bring improvement of the semantic maps.
The noticeable trend indicates that, with increasing number of the stacked units, the model benefits from the deeper network. Moreover, stacking multiple SDN units makes a coarse-to-fine prediction process, thus improving the network performance. However, it also leads to more computational stress and GPU memory demand, thus we stack up to 3 SDN units as more units bring too slight improvements.

\begin{table}[h]
\linespread{1.25}
\caption[Table1]{The performance comparison between different networks on PASCAL VOC 2012 val set.}
\begin{center}
\small
\begin{tabular}{c| c c c c c}
\toprule
   & {SDN${_M{_1}}$} & {SDN${_M{_2}}$}  & {SDN${_M{_3}}$} & {SDN${_M{_1}+}$} \\
\hline
\hline
  Depth & 169 & 185  & 201 & 185 \\
  Parameters (M)  &84.9  & 161.7 & 238.5 & 161.7 \\
  Mean IoU  & 78.2  & 79.2  & 79.9 & 78.6 \\
\bottomrule
\end{tabular}
\end{center}
\label{STACKT}
\end{table}

\subsubsection{Stacked network design}

From the experiments above, the network depth and its parameter size increase with the number of stacked units. We want to explore that the improvements are mainly brought by more parameters or stacked network design. To address this problem, we design a single encoder-decoder network, named as SDN${_M{_1}+}$, which has the same network depth and the size of parameters as the network stacking two SDN units, i.e., SDN${_M{_2}}$. SDN${_M{_1}+}$ also employ DenseNet-161 as the encoder module, meanwhile, the decoder module consists of 6 trivial convolutional layers and 2 upsampling blocks, the number of layers in upsampling block are set to 11 and 7 respectively.
The results for the performance comparison between SDN${_M{_1}+}$ and SDN${_M{_2}}$ are listed in Table \ref{STACKT}. From the results we can see that the performance of SDN${_M{_2}}$ is 0.6 percent higher than SDN${_M{_1}+}$. The comparison indicates that such a stacked network design, which captures more contextual information, improves the performance of the network.

\subsubsection{Hierarchical supervision}

To further alleviate the difficulty on optimization, we add hierarchical supervision to the upsampling
process in each unit.
In order to explore the impact of hierarchical supervision, we take SDN${_M{_1}}$ as an example to test the performance of different supervision, Specifically, we modify  SDN${_M{_1}}$ network by adding supervision at different upsampling scales, and refer corresponding settings to as SDN${_M{_{1\_1}}}$, SDN${_M{_{1\_2}}}$, SDN${_M{_1}}$ respectively. Here, we denote by \emph{up\_ratio} the ratio of the input image spatial resolution to the output resolution of the block. In SDN${_M{_1}}$ network, we add supervision at \emph{up\_ratio} {=} 16, 8, 4. And for SDN${_M{_{1\_2}}}$ network, we add supervision at \emph{up\_ratio} {=} 8, 4. In SDN${_M{_{1\_1}}}$ network, we only add supervision at \emph{up\_ratio} {=} 4. In the inference step, we output the prediction at different \emph{up\_ratio}.

Results are shown in Table \ref{STACKTTT}, we can find that, in SDN${_M{_1}}$ network, the performance at \emph{up\_ratio} {=} 4 is 2.3 percent higher than the performance at \emph{up\_ratio} {=} 16, it is clear that the high-resolution prediction, which use more low-level visual features, performs better than the low-resolution prediction during upsampling process of the unit.
Meanwhile, the performance keep increasing with the number of the supervision. the performance of SDN${_M{_{1\_2}}}$ is 0.5 percent higher than SDN${_M{_{1\_1}}}$, and there is modest improvement in performance from 78.0 to 78.2 when we add supervision at all \emph{up\_ratio}. The results prove the effectiveness of hierarchical supervision, and we employ hierarchical supervision during upsampling process of each unit.

\begin{table}[h]
\linespread{1.25}
\caption[Table3]{The performance comparison between different supervision on PASCAL VOC 2012 val set.}
\begin{center}
\small
\begin{tabular}{ c| c c c c c}
\toprule
& up\_ratio & {SDN${_M{_{1\_1}}}$} & {SDN${_M{_{1\_2}}}$} & {SDN${_M{_1}}$}  \\
\hline
\hline
& 16 & --  & -- & 75.9    \\

Mean IoU & 8 & --  &  77.1 & 77.7     \\

& 4 & 77.5  & 78.0 & 78.2  \\
\bottomrule

\end{tabular}
\end{center}
\label{STACKTTT}
\end{table}

We also explore the effects of score map fusion, depicted in Fig. 2. It is to compare the performance between networks with or without score map fusion. To this end, we stack two SDN units without score map connection, and refer to the corresponding network as SDN${_M{_{2}}-}$. The results are shown in Table \ref{STACKTT}, the performance of SDN${_M{_{2}}}$ is 0.4 percent higher than SDN${_M{_{2}}-}$. From the results we can see that the score map fusion is benefit for the segmentation result.

\begin{table}[h]
\linespread{1.25}
\caption[Table3]{The performance comparison networks with or without score map connections on PASCAL VOC 2012 val set.}
\begin{center}
\small
\begin{tabular}{ c| c c}
\toprule
 & {SDN${_M{_{2}}}$} & {SDN${_M{_{2}-}}$}   \\
\hline
\hline

Mean IoU  &  79.2 & 78.8   \\

\bottomrule

\end{tabular}
\end{center}
\label{STACKTT}
\end{table}

\subsubsection{Some improvement strategies}
In the section, detailed evaluations are performed on PASCAL VOC 2012 val dataset.
Here, we adopt several steps to improve segmentation performance further based on the SDN${_M{_2}}$ network: (1) UP: We further restore high resolution features by cascading a upsampling block, and we refer to the network as SDN${_M{_2}*}$. (2) MS\_Flip: We average the segmentation probability maps from 5 image scales \{0.5, 0.8, 1, 1.2, 1.4\} as well as their mirrors for inference. (3) COCO: Many state-of-the-art models adopt Microsoft COCO dataset \cite{MS_COCO} for the better performance. In order to compare with these models, we also pretrain the model of SDN${_M{_2}*}$ on MS-COCO dataset. We evaluate how each of these factors affects val set performance in Table \ref{STACKSTACK}.

\begin{table}[h]
\linespread{1.25}
\caption[Table3]{The performance comparison between different measures on PASCAL VOC 2012 val set.}
\begin{center}
\small
\begin{tabular}{c c c| c }
\toprule

 Up  & MS\_Flip &  COCO & Mean IoU \\
\hline
\hline
&    &  & 79.2 \\

  \checkmark  &   &   &   79.6  \\

  \checkmark   & \checkmark &   &   80.7  \\

  \checkmark  & \checkmark &  \checkmark &  84.8 \\

\bottomrule

\end{tabular}
\end{center}
\label{STACKSTACK}
\end{table}

Increasing an upsampling block gives 0.4\% gain, and segmentation map fusion brings another 1.2\% improvement. Moreover, when we pretrain the model of SDN${_M{_2}*}$ on MS-COCO dataset, the performance attains 84.8\%. Compared with current well-known method Deeplabv3 \cite{deeplabv3}(82.7\% pre-trained on MS-COCO), our method of the SDN${_M{_2}*}$ outperforms theirs by 2.1\%, which shows the strength of our proposed SDN network.
\subsubsection{Comparing with the state-of-the-art methods }
We further compare our method with the state-of-the-art methods on PASCAL VOC 2012 test set. Here, based on two settings, i.e., with or without pre-trained with MS-COCO dataset, we fine tune our the model of SDN${_M{_3}}$ on PASCAL VOC 2012 trainval set, and submit our test results to the official evaluation server. Results are shown in Table \ref{STACKVOC}. In the two settings, our method both outperforms all the other methods. Trained with only PSACAL VOC 2012 data, we achieve a Mean IoU score of 83.5\% \footnote{http://host.robots.ox.ac.uk:8080/anonymous/Z9RDVZ.html}.
When we pretrain the model of SDN${_M{_3}}$ on MS-COCO dataset, it reaches 86.6\% \footnote{http://host.robots.ox.ac.uk:8080/anonymous/GRWV3B.html} in Mean IoU.
Specifically, our model outperforms the RefineNet\cite{refinenet} by 2.4\%, and RefineNet employs deep encoder-decoder structure and demonstrates outstanding performance on several semantic segmentation dataset. Meanwhile, our model also outperforms the PSPNet\cite{refinenet} by 1.2\% and Deeplabv3 \cite{deeplabv3} by 0.9\%, they both adopt multi-scale or global features to capture contextual information. These comparisons indicate that our proposed SDN network can more effectively capture contextual information and generate accurate boundary localization.

\begin{table*}[t]
 \linespread{1.2}
\caption[Table1]{Experimental results on PASCAL VOC 2012 test set.}
\small
\setlength{\tabcolsep}{1.5pt}
\begin{center}
\begin{tabular}{ l |c c c c c c c c c c c c c c c c c c c c |c  }
\toprule
 Method & aero & bike& bird& boat& bottle &bus &car& cat& chair& cow& table& dog &horse& mbike &person &plant& sheep &sofa &train& tv& mIoU \\
\hline
 \multicolumn{22}{c}{\textbf{Only using VOC data}}\\
\hline
FCN\cite{FCN} & 76.8 &34.2& 68.9& 49.4& 60.3& 75.3& 74.7& 77.6& 21.4& 62.5& 46.8& 71.8 &63.9& 76.5& 73.9& 45.2& 72.4& 37.4& 70.9& 55.1& 62.2\\
DeepLab\cite{deeplabv2} &84.4&54.5&81.5&63.6&65.9&85.1&79.1&83.4&30.7&74.1&59.8&79.0&76.1&83.2&80.8&59.7&82.2&50.4&73.1&63.7&71.6\\
CRF-RNN\cite{CRF_RNN} &87.5&39.0&79.7&64.2&68.3&87.6&80.8&84.4&30.4&78.2&60.4&80.5&77.8&83.1&80.6&59.5&82.8&47.8&78.3&67.1&72.0\\
DeconvNet\cite{deconvnet} &89.9&39.3&79.7&63.9&68.2&87.4&81.2&86.1&28.5&77.0&62.0&79.0&80.3&83.6&80.2&58.8&83.4&54.3&80.7&65.0&72.5\\
GCRF\cite{GCRF} &85.2&43.9&83.3&65.2&68.3&89.0&82.7&85.3&31.1&79.5&63.3&80.5&79.3&85.5&81.0&60.5&85.5&52.0&77.3&65.1&73.2\\
DPN\cite{DPN} &87.7&59.4&78.4&64.9&70.3&89.3&83.5&86.1&31.7&79.9&62.6&81.9&80.0&83.5&82.3&60.5&83.2&53.4&77.9&65.0&74.1\\
Piecewise\cite{piecewise} &90.6&37.6&80.0&67.8&74.4&92.0&85.2&86.2&39.1&81.2&58.9&83.8&83.9&84.3&84.8&62.1&83.2&58.2&80.8&72.3&75.3\\
ResNet38\cite{resnet38} &94.4&72.9&\textbf{94.9}&68.8&\textbf{78.4}&90.6&\textbf{90.0}&92.1&40.1&90.4&71.7&89.9&93.7&\textbf{91.0}&89.1&71.3&90.7&61.3&87.7&\textbf{78.1}&82.5\\
PSPNet\cite{PSPNet} &91.8&71.9&94.7&71.2&75.8&95.2&89.9&95.9&39.3&90.7&71.7&90.5&\textbf{94.5}&88.8&89.6&\textbf{72.8}&89.6&\textbf{64.0}&85.1&76.3&82.6\\
\hline
SDN&\textbf{96.2}&\textbf{73.9}&94.0&\textbf{74.1}&76.1&\textbf{96.7}&89.9&\textbf{96.2}&\textbf{44.1}&\textbf{92.6}&\textbf{72.3}&\textbf{91.2}&94.1&89.2&\textbf{89.7}&71.2&\textbf{93.0}&59.0&\textbf{88.4}&76.5&\textbf{83.5}\\
\hline
\multicolumn{22}{c}{\textbf{Using VOC+COCO data}}\\
\hline
CRF-RNN\cite{CRF_RNN} &90.4&55.3&88.7&68.4&69.8&88.3&82.4&85.1&32.6&78.5&64.4&79.6&81.9&86.4&81.8&58.6&82.4&53.5&77.4&70.1&74.7\\
Dilation8\cite{yu2015multi} &91.7&39.6&87.8&63.1&71.8&89.7&82.9&89.8&37.2&84.0&63.0&83.3&89.0&83.8&85.1&56.8&87.6&56.0&80.2&64.7&75.3\\
DPN\cite{DPN} &89.0&61.6&87.7&66.8&74.7&91.2&84.3&87.6&36.5&86.3&66.1&84.4&87.8&85.6&85.4&63.6&87.3&61.3&79.4&66.4&77.5\\
Piecewise\cite{piecewise} &94.1&40.7&84.1&67.8&75.9&93.4&84.3&88.4&42.5&86.4&64.7&85.4&89.0&85.8&86.0&67.5&90.2&63.8&80.9&73.0&78.0\\
DeepLab\cite{deeplabv2} &92.6&60.4&91.6&63.4&76.3&95.0&88.4&92.6&32.7&88.5&67.6&89.6&92.1&87.0&87.4&63.3&88.3&60.0&86.8&74.5&79.7\\
RefineNet\cite{refinenet} &95.0&73.2&93.5&78.1&84.8&95.6&89.8&94.1&43.7&92.0&77.2&90.8&93.4&88.6&88.1&70.1&92.9&64.3&87.7&78.8&84.2\\
ResNet38\cite{resnet38} &96.2&75.2&95.4&74.4&81.7&93.7&89.9&92.5&48.2&92.0&79.9&90.1&95.5&91.8&91.2&73.0&90.5&65.4&88.7&80.6&84.9\\
PSPNet\cite{PSPNet} &95.8&72.7&95.0&78.9&84.4&94.7&\textbf{92.0}&95.7&43.1&91.0&\textbf{80.3}&91.3&96.3&\textbf{92.3}&90.1&71.5&\textbf{94.4}&66.9&88.8&82.0&85.4\\
DeepLabv3\cite{deeplabv3} & 96.4&76.6&92.7&77.8&\textbf{87.6}&96.7&90.2&95.4&47.5&93.4&76.3&91.4&\textbf{97.2}&91.0&\textbf{92.1}&71.3&90.9&\textbf{68.9}&\textbf{90.8}&79.3&85.7\\
\hline
SDN+&\textbf{96.9}&\textbf{78.6}&\textbf{96.0}&\textbf{79.6}&84.1&\textbf{97.1}&91.9&\textbf{96.6}&\textbf{48.5}&\textbf{94.3}&78.9&\textbf{93.6}&95.5&92.1&91.1&\textbf{75.0}&93.8&64.8&89.0&\textbf{84.6}&\textbf{86.6}\\
\bottomrule

\end{tabular}
\end{center}
\label{STACKVOC}
\end{table*}
\begin{figure*}[!t]
  \centering
  \centerline{\includegraphics[width = 1\linewidth]{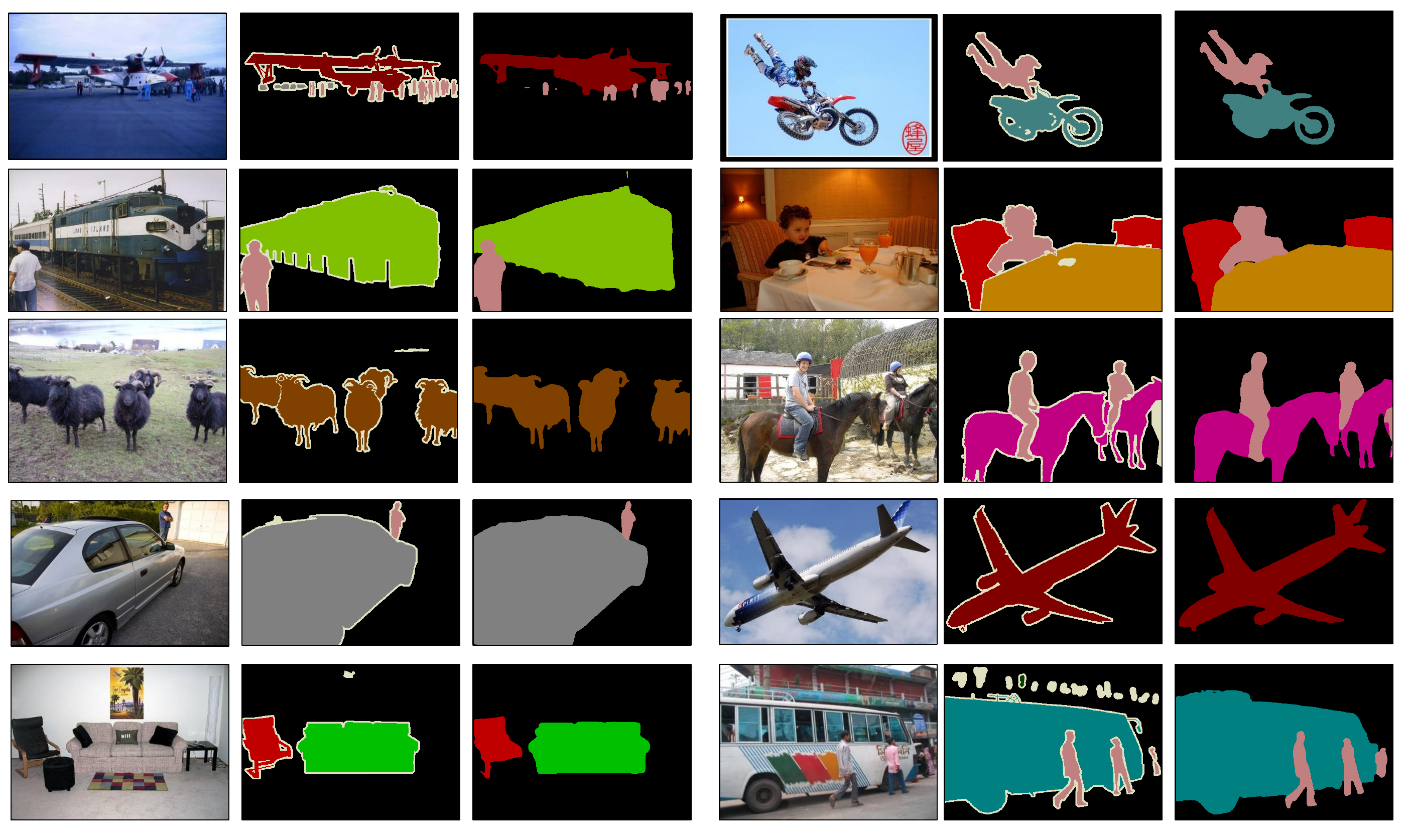}}
  \caption{Results on PASCAL VOC 2012 dataset. The images in each row from left to right are: (1) input image (2) groundtruth (3) semantic segmentation result.}
  \label{pascal_pic}
\end{figure*}

\subsection{Results on CamVid dataset}

\begin{figure*}[!t]
  \centering
  \centerline{\includegraphics[width = 1\linewidth]{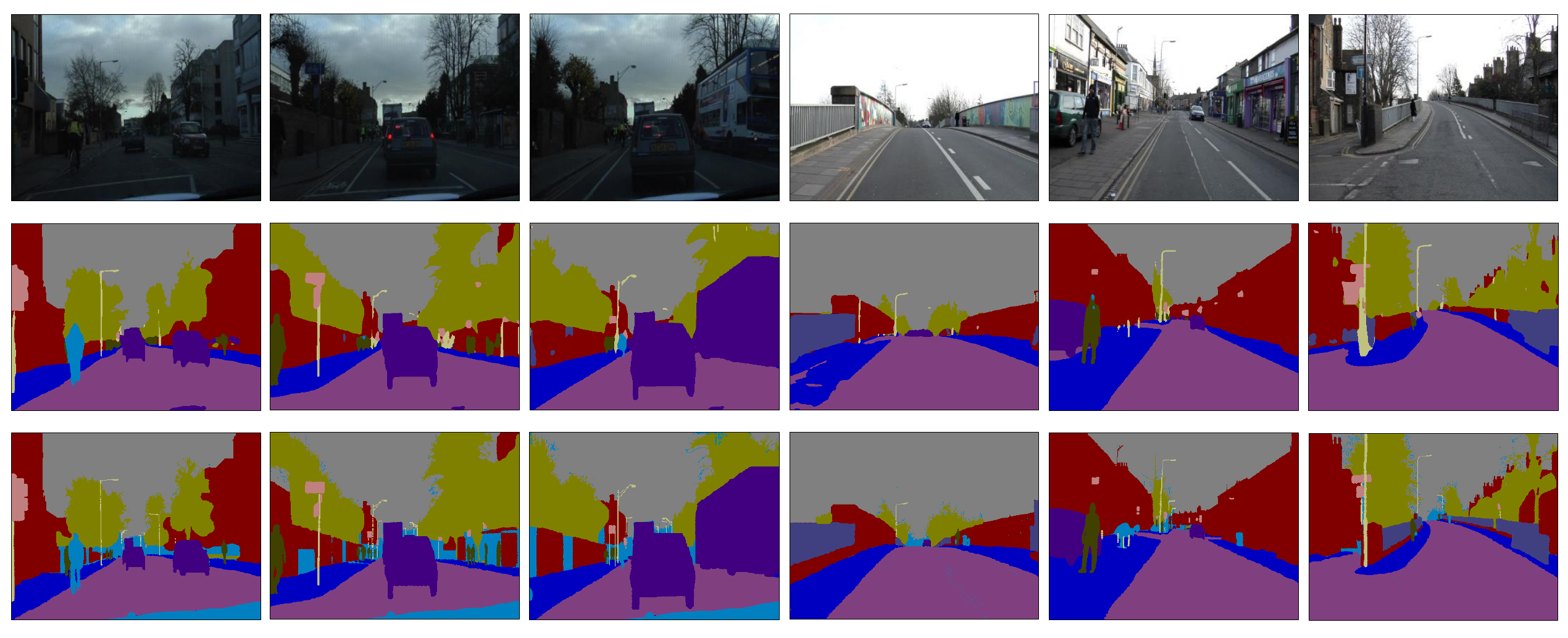}}
  \caption{Results on CamVid dataset. The images in each column from top to down are: (1) input image (2) semantic segmentation result (3) groundtruth.}
  \label{camvid_pic}
\end{figure*}

\begin{table*}[t]
\linespread{1.2}
\caption[Table1]{Experimental results on CamVid test set.}
\small
\setlength{\tabcolsep}{3.0pt}
\begin{center}
\begin{tabular}{ l |c c c c c c c c c c c c c c c c c c c c c c| c }
\toprule
 Method & Building & Tree& Sky& Car& Sign &Road &Pedestrian & Fence& Pole& Sidewalk& Bicyclist& Mean IoU& Global Avg \\
\hline
\hline
SegNet\cite{segnet} &68.7&52.0&87.0&58.5&13.4&86.2&25.3&17.9&16.0&60.5&24.8&46.4&62.5\\
DeconvNet\cite{deconvnet}&-- & --& --& --&-- &--& --&-- &-- & --&--  &48.9&85.6\\
ReSeg\cite{reseg} &-- & --& --& --&-- &--& --&-- &-- & --&--  &58.8 &88.7\\
DeepLab-LFOV\cite{fullyconnectedcrfs} &81.5&74.6&89.0&82.2&42.3&92.2&48.4&27.2&14.3&75.4&50.1&61.6&--\\
Bayesian SegNet\cite{bayes-segnet} &-- & --& --& --&-- &--& --&-- &-- & --&-- &63.1&86.9\\
Dilation8\cite{yu2015multi}  &82.6&76.2&89.0&84.0&46.9&92.2&56.3&35.8&23.4&75.3&55.5&65.3&79.0\\
HDCNN-448+TL \cite{wang2017hierarchically}  &-- & --& --& --&-- &--& --&-- &-- & --&--&65.6&90.9\\
Dilation8+FSO\cite{kundu2016feature} &84.0&77.2&91.3&85.6&49.9&92.5&59.1&37.6&16.9&76.0&57.2&66.1&88.3\\
FC-DenseNet103\cite{jegou2016one}&83.0&77.3&93.0&77.3&43.9&94.5&59.6&37.1&37.8&82.2&50.5&66.9&91.5\\
G-FRNet\cite{Refinement} &82.5&76.8&92.1&81.8&43.0&94.5&54.6&47.1&33.4&82.3&59.4&68.0&--\\
DCDN\cite{DCDN}&-- & --& --& --&-- &--& --&-- &-- & --&--&68.4&91.4\\
\hline
SDN&84.45&76.9&92.2&88.2&51.6&93.4&62.4&37.0&37.5&77.8&64.4&69.6&91.7\\
SDN+&\textbf{85.2}&\textbf{77.5}&92.3&\textbf{90.2}&\textbf{53.9}&\textbf{96.0}&\textbf{63.8}&39.8&\textbf{38.4}&\textbf{85.36}&\textbf{66.9}&\textbf{71.8}&\textbf{92.7}\\
\bottomrule

\end{tabular}
\end{center}
\label{STACKCamVid}
\end{table*}

In this subsection, we carry out experiments on the CamVid \cite{camvid} dataset to further evaluate the effectiveness of our method. The experimental settings are as follows: we apply SDN${_M{_2}*}$ network, and train the network with 367 training images, and test it with 233 test images. The initial learning rate is changed to 5e-5.
Meanwhile, we also adopt data augmentation to reduce overfitting by multi-scale translation.
Note that we compare SDN with the previous state-of-the-art methods on two settings, i.e., initializing the network with or without the model pretrained on PASCAL VOC 2012 \cite{voc}.

Following \cite{DCDN,jegou2016one,segnet}, Mean IoU and Global Avg are employed to evaluate our method on this dataset. We compare our method with previous ones in Table \ref{STACKCamVid}. Results show that our method on two settings both outperform other methods. The model, without pretrained on PASCAL VOC 2012(marked by SDN), achieves 69.6\% in Mean IoU and 91.7\% in Global Avg. Particularly, the classes, including \emph{Car}, \emph{Sign}, \emph{Pedestrian}, \emph{Bicyclist}, have a major boost in performance. When we initialize the network pretrained on PASCAL VOC dataset (marked by SDN+), the performance further improves by 2.2 percent in Mean IoU and 1 percent in Global Avg, and here nine out of eleven object categories achieve best performance in Mean IoU.
Note that, the CamVid dataset sampling in video frames contains temporal information, and some works \cite{kundu2016feature,wang2017hierarchically} mine temporal information to aid segmentation results. Our method still outperforms these works by a relative large margin. Besides, the spatio-temporal information of the dataset is complementary to our method and could bring additional improvements.

\begin{figure*}[!t]
  \centering
  \centerline{\includegraphics[width = 1\linewidth]{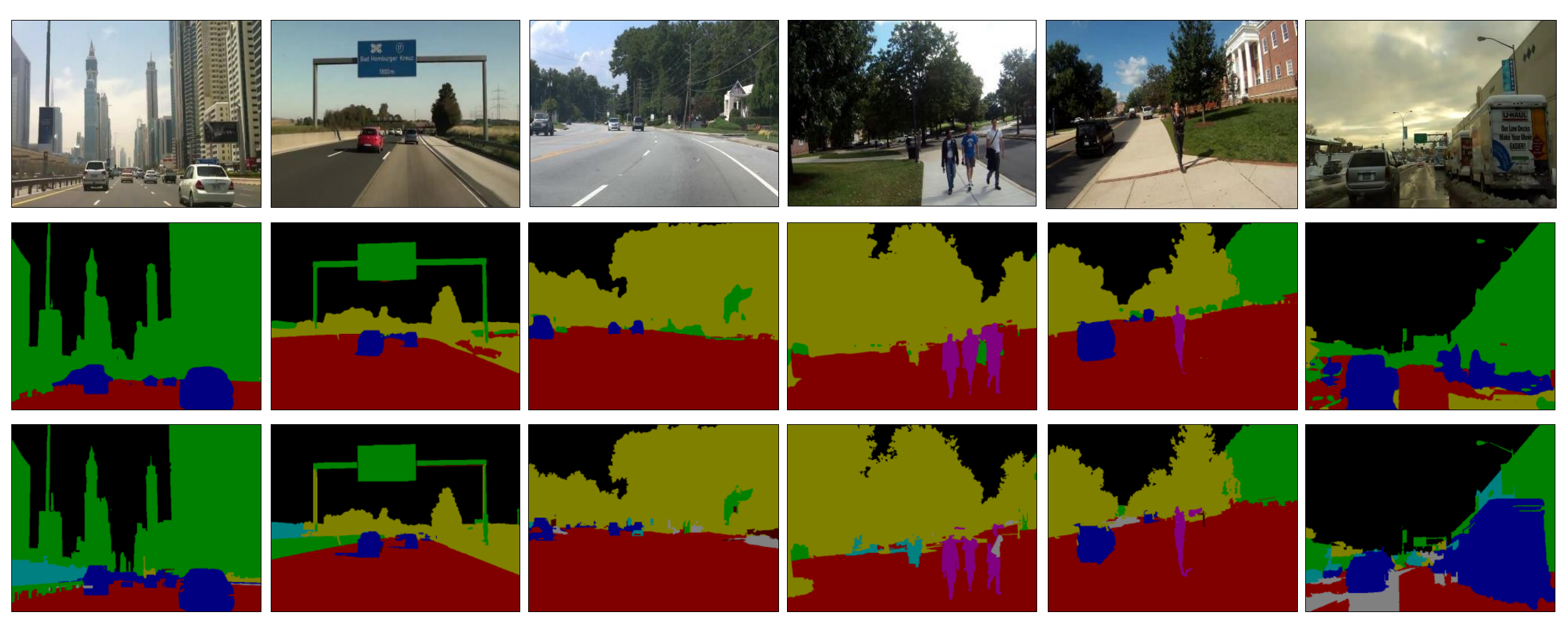}}
  \caption{Results on GATECH dataset. The images in each column from top to down are: (1) input image (2) semantic segmentation result (3) groundtruth.}
  \label{GATECH_pic}
\end{figure*}

Some test images along with ground truth and our predicted semantic maps are shown in Fig. \ref{camvid_pic}. We can find that
our network can well sketch multi-scale appearance of objects, including large-scale  objects,i.e., building and car, and shape objects i.e., poles and pedestrians. All the results on the this dataset show that our network can capture more contextual information and learn better spatial-relationship.

\subsection{Results on GATECH dataset}

In order to further verify the generalization of our models, we evaluate our network on GATECH \cite{GATECH} dataset, which is much larger than CamVid dataset, but has a lot of noisy annotations. We employ the SDN${_M{_2}*}$ network with the same training strategy on CamVid. We also compare our models with previous state-of-the-art methods on two settings, i.e., initializing the network with (marked by SDN) or without (marked by SDN+) the model pretrained on PASCAL VOC 2012 \cite{voc}.

\begin{table}[h]
\linespread{1.2}
\caption[Table2]{Experimental results on GATECH test set.}
\small
\setlength{\tabcolsep}{3.0pt}
\begin{center}
\begin{tabular}{c|c c c}
\toprule
  & Temporal & Global & Mean  \\
Method  &  Info &  Avg &  IoU \\
\hline
\hline
3D-V2V-scratch  \cite{tran2016deep} & Yes &66.7&- \\

3D-V2V-finetune \cite{tran2016deep} & Yes & 76.0&- \\

FC-DenseNet103 \cite{jegou2016one} & No & 79.4 & - \\

HDCNN-448+TL \cite{wang2017hierarchically} & Yes & 82.1  &  48.2 \\
DCDN\cite{DCDN} & No & 83.5 & 49.0\\
\hline
SDN & No & 84.6 & 53.5\\
SDN+ & No & \textbf{86.3} & \textbf{55.9}\\
\bottomrule
\end{tabular}
\end{center}

\label{GATECH}
\end{table}

Results are shown in Table \ref{GATECH}, we can find that our method on two settings both outperforms other methods.
SDN yields 53.5\% in mean IoU and 84.6\% in Global Avg. and SDN+ improves the result significantly, which gives 1.7\% gain in Global Avg and 2.4\% in Mean IoU. Specially, our model, without using any temporal, performs better than the models \cite{tran2016deep,wang2017hierarchically} which exploit spatio-temporal relationships between video frames. All these comparisons confirm the proposed SDN a robust and effective model for coarse-annotation dataset. Some test images along with ground truth and our predicted semantic maps are shown in Fig. \ref{GATECH_pic}.

\section{Conclusion}

In this paper, we have presented a stacked deconvolutional network (SDN), a novel deep network architecture for semantic segmentation. We stack multiple SDN units to make network deeper and realize a coarse-to-fine learning process, meanwhile, intra-unit and inter-unit connections and hierarchical supervision are adopted to promote network optimization. The ablation experiments show that those designs effectively capture contextual information and recover the spatial resolution for accurate boundary localization, which benefit network performance. Our best model outperform all previous works on three public benchmarks.

\bibliographystyle{IEEEtran}
\bibliography{sigproc_bib}

\end{document}